\pdfoutput=1

\documentclass[11pt]{article}

\usepackage[final]{coling} 

\usepackage{times}
\usepackage{latexsym}

\usepackage[T1]{fontenc}

\usepackage[utf8]{inputenc}

\usepackage{microtype}

\usepackage{inconsolata}

\usepackage{graphicx}

\usepackage{amsmath} 
\usepackage{multirow} %
\usepackage{float}
\usepackage{subfig}   %
\usepackage{makecell}
\usepackage{tabularx}
\usepackage{multicol}
\usepackage{booktabs} 

\usepackage{amssymb}
\usepackage{bbding}
\usepackage{pifont}
\usepackage{wasysym}
\usepackage{utfsym}
\usepackage{fontawesome}
\newcommand{\cmark}{\ding{51}}%
\newcommand{\xmark}{\ding{55}}%

\usepackage[marginal]{footmisc}

\newcommand{\ourmethod}{\textsc{XFormParser}}

\title{\ourmethod{}: A Simple and Effective Multimodal Multilingual Semi-structured Form Parser}

\author{
 \textbf{Xianfu Cheng\textsuperscript{1\dag}},
 \textbf{Hang Zhang\textsuperscript{1\dag}}, 
 \textbf{Jian Yang\textsuperscript{2}},
 \textbf{Xiang Li\textsuperscript{2}},
 \textbf{Weixiao Zhou\textsuperscript{1}},
 \textbf{Fei Liu\textsuperscript{3}},
 \textbf{Kui Wu\textsuperscript{1}},
 \\
 \textbf{Xiangyuan Guan\textsuperscript{1}},
 \textbf{Tao Sun\textsuperscript{1}},
 \textbf{Xianjie Wu\textsuperscript{1}},
 \textbf{Tongliang Li\textsuperscript{4*}},
 \textbf{Zhoujun Li\textsuperscript{1,5*}},
\\
 \textsuperscript{1}CCSE, Beihang University,
 \textsuperscript{2}Beihang University,
 \textsuperscript{3}Beijing Language and Culture University,
 \\
 \textsuperscript{4}Beijing Information Science and Technology University,
 \\
 \textsuperscript{5}Shenzhen Intelligent Strong Technology Co.,Ltd.
\\
 \{buaacxf, zhbuaa0, jiaya, xlggg, wxzhou, gxy0615, buaast, wuxianjie, lizj\}@buaa.edu.cn,
 \\
 17861517116@163.com, wukui0099@gmail.com, tonyliangli@bistu.edu.cn
}

\begin{document}
\maketitle

\footnote{\dag Both authors contributed equally to this research.}
\footnote{* Corresponding Author.}
\footnote{$^1$ The codes, datasets, and pre-trained models are publicly available at https://github.com/zhbuaa0/xformparser.}

\begin{abstract}
In the domain of Document AI, parsing semi-structured image form is a crucial Key Information Extraction (KIE) task. The advent of pre-trained multimodal models significantly empowers Document AI frameworks to extract key information from form documents in different formats such as PDF, Word, and images. Nonetheless, form parsing is still encumbered by notable challenges like subpar capabilities in multilingual parsing and diminished recall in industrial contexts in rich text and rich visuals. In this work, we introduce a simple but effective \textbf{M}ultimodal and \textbf{M}ultilingual semi-structured \textbf{FORM} \textbf{PARSER} (\textbf{XFormParser}), which anchored on a comprehensive Transformer-based pre-trained language model and innovatively amalgamates semantic entity recognition (SER) and relation extraction (RE) into a unified framework. Combined with Bi-LSTM, the performance of multilingual parsing is significantly improved.
Furthermore, we develop InDFormSFT, a pioneering supervised fine-tuning (SFT) industrial dataset that specifically addresses the parsing needs of forms in various industrial contexts. XFormParser has demonstrated its unparalleled effectiveness and robustness through rigorous testing on established benchmarks. 
Compared to existing state-of-the-art (SOTA) models, XFormParser notably achieves up to 1.79\% F1 score improvement on RE tasks in language-specific settings.  It also exhibits exceptional cross-task performance improvements in multilingual and zero-shot settings.$^1$ 

\end{abstract}

\section{Introduction}
\begin{figure}[!t]
	\centering
	\includegraphics[width=1.0\linewidth]{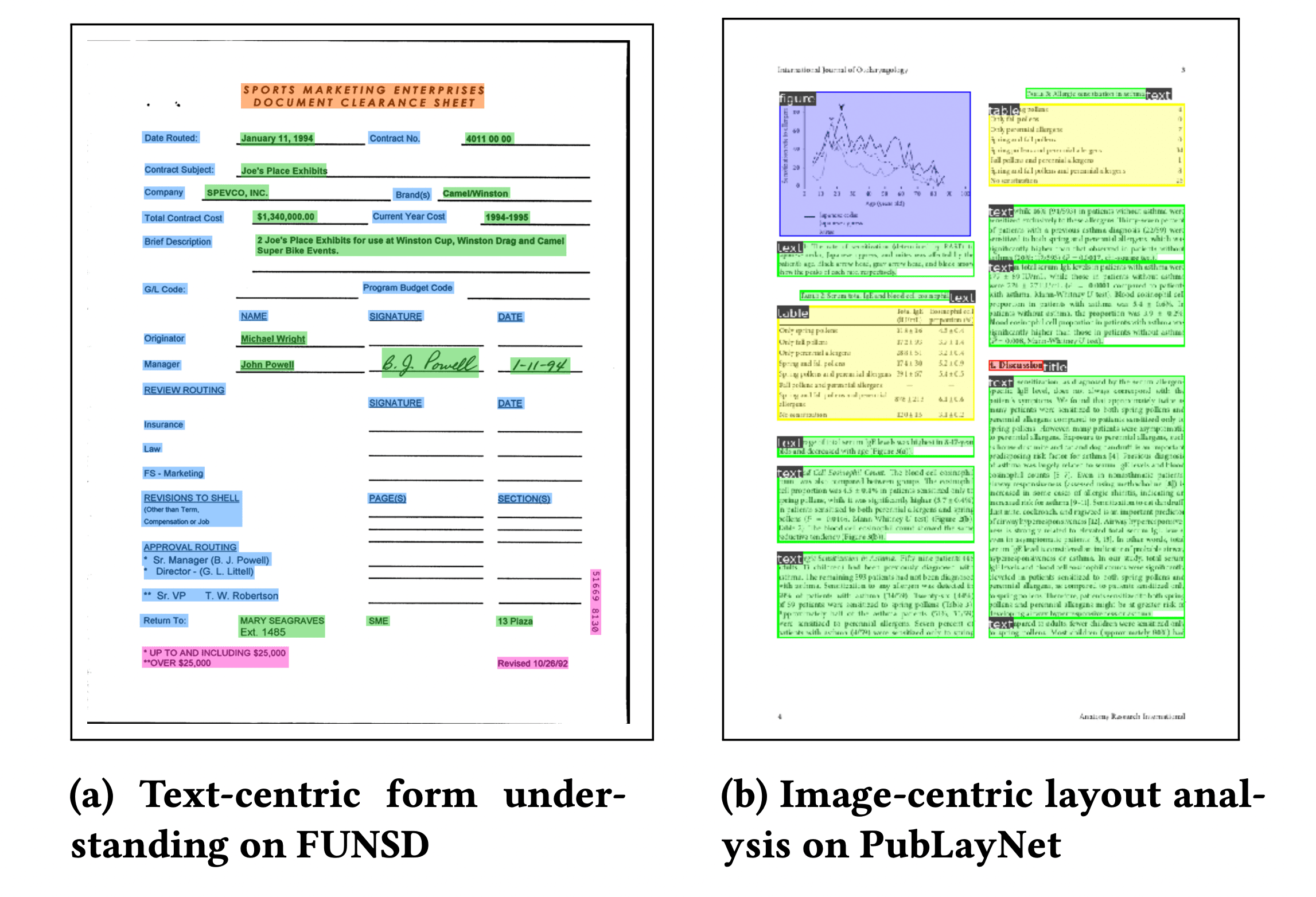}
	\caption{An illustration of named entity recognition for unstructured forms.}
	\label{fig:formner}
\end{figure}

Document AI is the technology of automatically reading, understanding, and analyzing business documents. It is widely applied in commercial, governmental, and educational sectors and is crucial to departmental efficiency and productivity. A key task of Document AI is to parse and extract form information from scanned documents. As shown in Figure~\ref{fig:formner}, form parsing is essentially an entity relation mining task that connects the Named Entity Recognition (NER) task \cite{li2020survey} and KIE task \cite{cui2021document, yu2021pick, hong2022bros}. Due to the diversity of the layout and the format, poor quality of scanned document images, and complexity of template structures, representing and understanding the unstructured information in documents using generic rules becomes a highly challenging task.

Unlike traditional NER tasks which only deal with textual information, and traditional pattern recognition tasks \cite{medvet2011probabilistic,cheng2020improved}, mainstream Document AI methods typically involve using deep neural networks to model elements in documents from the perspectives of computer vision (CV)~\cite{he2016deep,ren2016faster}, natural language processing (NLP) \cite{zhou2023multi,zhang2024mabcmultiagentblockchaininspiredcollaboration}, or multimodal fusion \cite{li2024adaptive}. Apart from textual information, the position and layout of text blocks also play a crucial role in semantic interpretation. 

Form parsing algorithms based on deep learning initially involved detecting and classifying specific regions of document images. Then, it utilized Optical Character Recognition (OCR) models \cite{li2023trocr,cheng2024viptr} to extract text information. This process is called unstructured to semi-structured, and the resulting form is called semi-structured. Subsequently, different categories of form blocks and their corresponding text content were routed to dedicated information extraction modules, constructing a pipeline to process entire-page documents \cite{2020LayoutLMv2, wang2022lilt}.
Current researchers believe that effective language models for form parsing must comprehend target entities and adapt to different document formats in contexts involving multiple modalities. Advanced Document AI models~\cite{wang2020docstruct,zhang2020trie,li2021structurallm,peng2022ernie} are expected to automatically classify, extract, and structure information from business documents, minimizing manual intervention.

Although advanced models have made significant progress in the field of Semi-structured form parsing, as research and applications have become more widespread, most existing methods still face three limitations: 1) The accuracy of key information extraction from complex, multilingual form images remains insufficient; 2) Multimodal models dominated by visual modalities still lag behind text-dominated multimodal models in form parsing tasks in rich-text and long-text scenarios; 3) Multi-modal large language models (MLLM) such as GPT4o~\cite{huang2023chatgpt,islam2024gpt} and LayoutLLM~\cite{fujitake2024layoutllm} are difficult to deploy to the end side and achieve fast and high-performance inference via CPUs or low-memory GPUs due to the excessive weight of model parameters. Therefore, in industrial applications, it is a crucial research direction to further mine the entity classification and the relations between entities in multimodal and multilingual forms based on simple and effective pre-trained models (PTM)~\cite{zoph2020rethinking}, and to study more effective fine-tuning paradigms in complex scenes.

To address these issues, we propose a simple but effective semi-structured form parser with multimodal and multilingual knowledge, named XFormParser. For input data from semi-structured forms, XFormParser utilizes the multilingual document understanding PTM LayoutXLM~\cite{xu-etal-2022-xfund} to generate vectors containing text, visual, and spatial positional information. Subsequently, these vectors are fed into the downstream joint network to complete two tasks: Semantic Entity Recognition (SER) and Relation Extraction (RE), to realize form parsing. The SER Task obtains text box classification through fully connected layers, and the RE task learns the categories of entity relations through a decoder based on Bi-LSTM \cite{sun2022neural} and Biaffine \cite{nguyen2019end}. In addition, we further build InDFormSFT, a Chinese and English multi-scenario form parsing SFT dataset for industrial applications, based on public benchmarks. Training the model on this dataset helps XFormParser learn semi-structured forms from the real world and achieve new SOTA performance.

The contributions of this paper are summarized as follows:
\begin{itemize}
\item We propose XFormParser, which integrates two tasks: SER and RE, along with the joint loss function and training method of soft labels warm-up in stages. XFormParser effectively enhances form parsing performance without additional inference resources and overhead.


\item We construct InDFormSFT, a cross-scenario form parsing SFT dataset in both Chinese and English. It contains 562 form images collected from 8 major industrial application scenarios and corresponding annotation information in JSON format that is semi-automatically generated using tools such as GPT4o and rigorously verified by humans.


\item Through experiments and analysis, we confirm that XFormParser achieves an F1 score of at most 1.79\% over the SOTA model for RE tasks in Language-specific scenarios. XFormParser achieves significantly better results than SOTA for both dual-task in Multi-language and Zero-shot scenarios. The ablation experiments on InDFormSFT demonstrate the effectiveness and robustness of XFormParser.

\end{itemize}

\begin{figure*}[!h]
	\centering
	\includegraphics[width=0.95\textwidth]{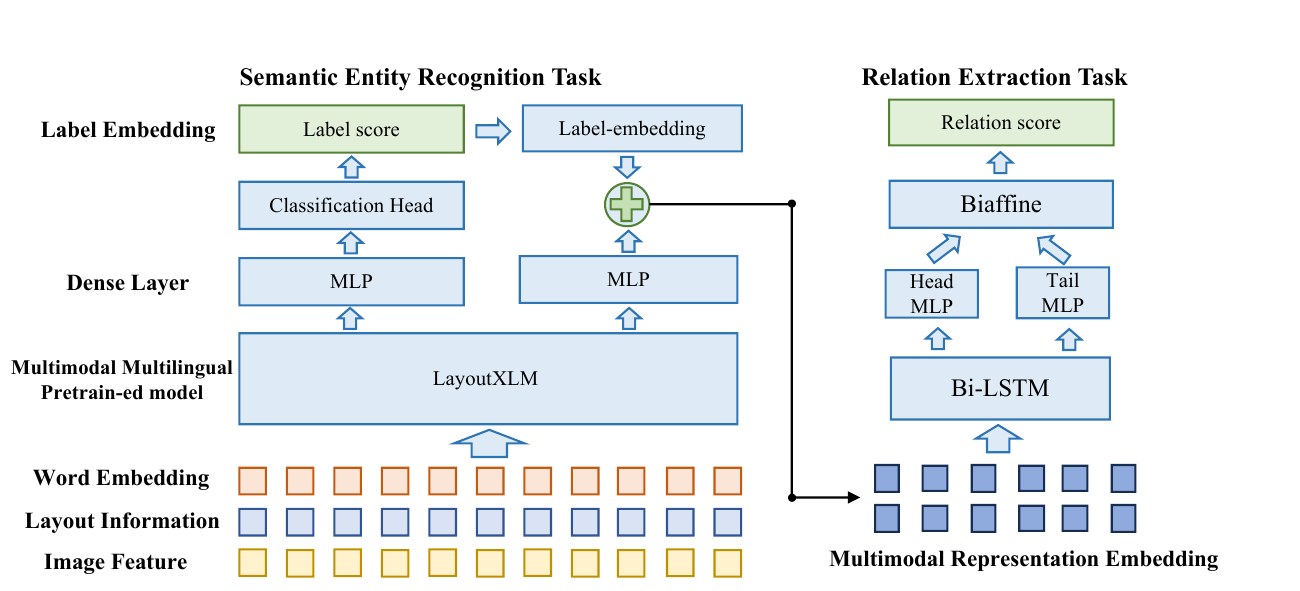}
	\caption{Overall architecture of the proposed XFormParser. For Multimodal input, XFormParser utilizes layoutXLM to generate vectors containing text, visual, and spatial positional information. Subsequently, these vectors are fed into the downstream joint network to complete SER and RE tasks. The SER Task obtains text box classification through fully connected layers, and the RE task learns the categories of entity relations through a decoder based on Bi-LSTM and Biaffine.}
	\label{fig:XFormParser}
    \vspace{-5pt}
\end{figure*}

\section{Related Work}
Limited by the lack of training data and the complexity of the corpus, relatively few effective models for parsing multilingual forms have been proposed in the past few years. Most applications use a pipeline approach to process form information one modality by one, such as XLM-RoBERTa~\cite{conneau2019unsupervised}, which is a multilingual version of RoBERTa~\cite{liu2019roberta}, and InfoXLM~\cite{chi2021infoxlm}, a multilingual pre-trained model~\cite{transformer,openai2023gpt4} that maximizes the mutual information between multilingual and multi-granularity texts, which are used as text information processors to analyze forms.

The Visual-Language models \cite{wu2021cvt, radford2019language,zhang2024lemurlogparsingentropy} demonstrate the potential to effectively align gaps between textual and other modal features, which can be used to integrate form structure information and content information. It is further revealed that form information extraction can be achieved by multi-modal techniques combining CV \cite{cheng2022improved} and NLP~\cite{m3p,soft_template,hlt_mt,low_resource_template,xcot,zhang2024eclipsesemanticentropylcscrosslingual}.

The LiLT \cite{wang2022lilt} model proposes a language-independent transformer focusing on text layout, introducing a new Bidirectional Attention Complement Mechanism to enhance cross-modal cooperation. It also proposes Key Point Localization and Cross-modal Alignment Identification tasks, combined with the widely used Masked Visual Language Model as pre-training objectives.
LayoutLM \cite{2020LayoutLM} introduced the PTM, which combined language models and Transformer, expanding BERT~\cite{devlin2018bert} architecture by incorporating layout information to consider spatial relations between tokens and textual content connections, resulting in outstanding performance in form parsing tasks. Furthermore, LayoutLMv2 \cite{2020LayoutLMv2} and LayoutLMv3 \cite{huang2022layoutlmv3} enhance the integration of multimodal information in the pretraining stage. They also add pretraining strategies for text-image alignment and text-image matching, incorporating word token alignment targets. 



Regarding the joint modeling of SER and RE, existing work primarily targets sequence labeling tasks, training jointly at the token granularity \cite{jiang2024entity,ji2024span,wang2023deep,el2024information,liu2023centre}. At present, there is a lack of available models or methods for joint pattern training with cell information.

GeoLayoutLM~\cite{luo2023geolayoutlm} improves the feature representation of text and layout by explicitly modeling geometric relationships and special pre-training tasks, which can improve the performance of form information extraction. However, the complex model structure makes the pre-training cost of related tasks high and too dependent on data.
GOSE \cite{chen2023global} first generates initial relation predictions for entity pairs extracted from document scan images. It then captures global structure knowledge from previous iterative predictions and incorporates it into entity representations. This "generate-capture-incorporate" loop is repeated multiple times, allowing entity representations and global structure knowledge to reinforce each other.

\section{Method}
\subsection{Task Definition}
The semi-structured form processed by the OCR model can be represented as a list of $n$ semantic entities. Each entity consists of a set of words named $Words_i$, the coordinates of the bounding box named $Bbox_i$, and an image named $Image_i$, called $i$-th cell of the form and defined as $b_i$:
\begin{equation}
  b_i = [Words_i, Bbox_i, Image_i] 
\end{equation}
The documents in our dataset are annotated with labels for each entity and relations between entities. We represent each comment document $D$ as:

\begin{equation}
  D=[B, L, R]
\end{equation}

where $B$=$[b_1,\ldots,b_n]$ denote all the cells of the form; $L$=$[{l_1,\ldots,l_n}]$ is a predefined set of entity labels and $l$ is the label of each entity; $R$=$[{(b_1, b_2),\ldots,(b_j, b_k)}]$ is the set of relations between entities, $(b_j, b_k)$ refers to the relation between $j$-th entity and $k$-th entity, as well as the link point from $b_k$ to $b_j$. It is worth noting that one entity may have relations with multiple entities or may not have a relationship with some other entity.

\textbf{Semantic Entity Recognition Task.} The semantic entity recognition needs to classify all the cells and obtain a list of resulting labels $L$, the label of cells can be: Header entity (HEADER), QUESTION entity (QUESTION), ANSWER entity (ANSWER) and OTHER entity (OTHER). 
When the entity classification is completed, each token needs to be converted into a BIO label according to the entity category, to facilitate the alignment with experiments on open-source benchmark datasets.

\textbf{Relation Extraction Task.} Given the above definition and description of a form, for a form $D$, each cell $b_i$ needs to find its corresponding sequence of cells in the whole form $B_{b_i}$=$[{b_j,\ldots,b_k}]$, Finally, the normalized cell relation $R_{b_i}$=$[\{(b_i,b_j),\ldots, (b_i, b_k)\}]$ is obtained.

\subsection{Overall Architecture}
As shown in Figure \ref{fig:XFormParser}, we built the XFormParser model based on the multimodal Transformer architecture. The model accepts information from three different modalities: text, position, and vision, encoded using text embedding, 2D position embedding, and image embedding layers, respectively. Concatenate the text and image embeddings and then add the position embeddings to obtain the input embeddings. The input embedding is encoded by layoutXLM, and the context representation is output through the dense layer. On the one hand, the representation vector is passed through the MLP layer to obtain the entity classification. On the other hand, the entity embedding vector is mined through the Bi-LSTM sequence relation and then entered into the $MLP_{head}$ or $MLP_{tail}$ according to the type to obtain the entity expression. The $Score$ is obtained by Biaffine to determine whether there is a relation between two entities.

\subsection{PTM and Multimodal Input}
layoutXLM adds two new embedding layers, 2-D Position Embedding and Image Embedding, based on BERT. The 2-D Position Embedding corresponds to text blocks with the content and position information in the document, both content and position information obtained by OCR and other technologies. By considering the top-left corner of the document page as the origin of the coordinate system, the representation of each text block in terms of horizontal coordinate, vertical coordinate, width, and height can be calculated, and the final 2-D Position Embedding is the sum of the representations of the four sub-layers. The image embedding divides the image into several blocks based on the bounding boxes of each word in the OCR results, and each block corresponds to each word. After normalizing embeddings of multiple modalities, they are input into layoutXLM.

\textbf{Vectorization of Text Information.} 
Tokenization, word embeddings, and sentence embeddings mainly realize the vectorization of text information. Among them, tokenization breaks down text into smaller units such as words or subwords.
LayoutXLM uses RoBERTa Tokenizers to map these tokens into dense vectors in a continuous vector space where semantically similar words are positioned closely together. Sentence embeddings are also used, provide a vector representation for entire sentences, capturing the semantic meaning of the sentence as a whole. 

\textbf{Vectorization of Location Information.}
Two-dimensional position embedding uses OCR and other techniques to obtain the content and position information of each text block in the document. The upper-left corner of the document page is regarded as the origin of constructing the coordinate system so that the embeddings of each text block corresponding to XY coordinates, width, and height can be calculated, and the final two-dimensional position embedding is the sum of the four embeddings.

\textbf{Vectorization of Image Feature.}
In this work, the image embedding is not simply obtained by uniform segmentation, but the text block is used to locate the bounding box of each word in the result, and the image is divided into several equal size patches, to ensure that each word has embedding vector of the image containing it.

\subsection{Semantic Entity Recognizer}
As shown in Figure \ref{fig:XFormParser}, the PTM receives multi-modal input $B$=$[{b_1,\ldots, b_n}]$, obtains hidden states $H$, and processes the hidden representation $H$ through a fully connected layer (Dense). The hidden representation $H_{ser}$ for the SER task is obtained, and then the classification result $logits_{ser}$ is obtained by a classification MLP. The formulas are expressed as follows: 
\begin{equation}
    H = PLM(B)
\end{equation}
\begin{equation}\label{eq:dense}
    H_{ser} = Dense_{ser}(H) 
\end{equation}
\begin{equation}
    logits_{ser} = MLP_{ser}(H_{ser}) 
\end{equation}
Then $logits_{ser}$ uses $argmax$ to get the predicted vectors $P$=$[p_1,\ldots, p_n]$, $p_i$ corresponds to the label $l_i$ of $i$-th entity, such as "QUESTION", "ANSWER", etc. 

\subsection{Relation Extraction Decoder}
For the RE task, we first get the PTM output of the $i$-th entity, denoted as $H_{re}$. The model then uses a pooling operation~\cite{reimers2019sentence} on $H_{re}$ to obtain the embeddings of the entities. Then we concatenate the entity embeddings and the corresponding label embeddings $p_i$ predicted by the SER. As the output of the encoder for each semantic entity, the formula is as follows:
\begin{equation}\label{eq:encoder}
    e_i = pooling(H_{re})\oplus p_i 
\end{equation}
Note that for two tasks to implicitly share $H$, $H_{re}$ needs to be obtained using a Dense layer identical to Eq.~\eqref{eq:dense}, that is, $H_{re}$= $Dense_{re}(H)$. In addition, mean-pooling is experimentally verified to be the most effective pooling operation applied in Eq.~\eqref{eq:encoder}.

In the experiment, it was found that the distribution of entity expression and label embedding was not uniform, which would make it difficult for MLP to learn the importance of both information. Therefore, $e_i$ was input into the Bi-LSTM decoder to unify, and after the entity passed the decoder, according to the type, it entered the head entity decoder $MLP_{head}$ or the tail entity decoder $MLP_{tail}$, and finally produces the entity representation. It obtains the score through Biaffine to determine whether there is a relation between the two entities.

\subsection{Traning Method}
\textbf{Training Loss.}
Since it is a joint model, the team must be trained to calculate the loss. The loss formula is as follows:
\begin{equation}
    Loss_{ser} = \text{CE}(logits_{ser},L_{ser})
\end{equation}
\begin{equation}
    Loss_{re} = \text{CE}(Score_{Biaffine},L_{re})
\end{equation}
\begin{equation}
    Loss = Loss_{ser} + Loss_{re}
\end{equation}
where \text{CE} is the cross-entropy function. $Loss_{ser}$ is the loss for the SER task, $Loss_{re}$ is the loss for the RE task, and the final training Loss is the sum of the two used as the learning target of the joint task. 
 
\textbf{Warm-up Soft Label.} We propose an improved warm-up soft label~\cite{huang2019bertbased} based on the warming mechanism, which can effectively improve the performance when applied to fine-tuning form parsing models. In the beginning stage of training, hard labels are used to supervise the model so that it can converge quickly. In the middle and later stages of training, soft labels are used to help model training. During the transition period between soft and hard labels, a warming mechanism is added, and the weight of soft tags is continuously increased. This is useful for tasks such as multi-label classification. 

In actual experiments, it was found that the training effect was not good in the first half of model training. Although soft labels contain more information, they cannot guide the model in learning tasks in the early stage of training, so hard labels are still used in the early stage of model training. After the model has been trained to have preliminary capabilities, then use soft label training, and provide a transition for the conversion of hard label and soft label. For the construction method of the warm-up soft label and related training parameter Settings, please refer to Appendix~\ref{sec:softlabel}.

\begin{table}[!h]
	\centering
	\begin{tabular}{c|c}
		\toprule
		Setup & Multi-language model   \\
		\midrule
		Optimizer &{AdamW}                          \\
		Weight ratio &{0.1}                         \\
		Lr scheduler &{LINEAR}                         \\
		vocab size & 250002                        \\
		Max Steps  & 512                         \\
		Batch size  & 8                         \\
		Initial learning rate &5e-5 \\
		Training epochs  &100                         \\
		Evaluation metric & ref. \ref{app:b4}                       \\
		\bottomrule
	\end{tabular}
        \caption{Training configurations for XFormParser}
	\label{tab:config}
\end{table}

\begin{table*}[!h]
    \fontsize{8}{8}\selectfont
    \centering
    \begin{tabular}{c|cccccccccc} 
    \toprule

    &Model&FUNSD&ZH&JA&ES&FR&IT&DE&PT&Avg.\\
    \midrule
    \multirow{7}*{SER}&XLM-RoBERTa\textsubscript{BASE}&66.70&87.74&77.61&61.05&67.43&66.87&68.14&68.18&70.47\\
    &InfoXLM\textsubscript{BASE}&68.52&88.68&78.65&62.30&70.15&67.51&70.63&70.08&72.07\\
    &LayoutXLM\textsubscript{BASE}&79.40&89.24&79.21&75.50&79.02&80.82&82.22&79.03&80.56\\
    &LiLT[InfoXLM]\textsubscript{BASE}&84.15&89.38&79.64&79.11&79.53&83.76&82.31&82.20&82.51\\
    &GeoLayoutLM&92.86&-& -& -& -& -& -& -& -\\    
    &XFormParser[LiLT]&91.42&91.89&82.25&87.18&87.64&89.42&87.05&87.86&88.09\textsubscript{{\color{red} (+5.58)}}\\
    &XFormParser&\textbf{92.46}&\textbf{93.14}&\textbf{82.59}&\textbf{87.77}&\textbf{88.69}&\textbf{90.51}&\textbf{88.48}&\textbf{88.68}&\textbf{89.04}\textsubscript{{\color{red} (+6.53)}}\\
    \cmidrule{1-11}
    \multirow{8}*{RE}
    &XLM-RoBERTa\textsubscript{BASE}   &26.59&51.05&58.00&52.95&49.65&53.05&50.41&39.82&47.69\\
    &InfoXLM\textsubscript{BASE}       &29.20&52.14&60.00&55.16&49.13&52.81&52.62&41.70&49.10\\
    &LayoutXLM\textsubscript{BASE}     &54.83&70.73&69.63&68.96&63.53&64.15&65.51&57.18&64.32\\
    &LiLT[InfoXLM]\textsubscript{BASE} &62.76&72.97&70.37&71.95&69.65&70.43&65.58&58.74&67.81\\
    &GeoLayoutLM                       &89.45& -& -& -& -& -& -& -& -\\
    &GOSE[LiLT]&76.97&87.52&80.96&85.95&86.46&84.15&80.23&73.84&82.01 \\
    &XFormParser[LiLT]&90.02&92.00&91.32&90.21&91.01&91.37&91.11&88.48&90.82\textsubscript{{\color{red} (+8.81)}} \\
    &XFormParser&\textbf{91.24}&\textbf{93.42}&\textbf{92.19}&\textbf{90.82}&\textbf{91.55}&\textbf{92.48}&\textbf{92.36}&\textbf{89.12}&\textbf{91.65}\textsubscript{{\color{red} (+9.64)}}\\
    \bottomrule
    \end{tabular}
    \caption{Language-specific fine-tuning F1 accuracy on FUNSD and XFUND (fine-tuning on X, testing on X).“SER” denotes the semantic entity recognition, and “RE” denotes the relation extraction.}
    \label{tab:result_xfund_specific}
\end{table*}

\begin{table*}[!h]
    \fontsize{8}{8}\selectfont
    \centering
    \begin{tabular}{c|cccccccccc} 
    \toprule

    &Model&FUNSD&ZH&JA&ES&FR&IT&DE&PT&Avg.\\
    \midrule
    \multirow{5}*{SER} & XLM-RoBERTa\textsubscript{BASE}&66.33&88.3&77.86&62.23&70.35&68.14&71.46&67.26&71.49\\
    & InfoXLM\textsubscript{BASE}&65.38&87.41&78.55&59.79&70.57&68.26&70.55&67.96&71.06 \\
    & LayoutXLM\textsubscript{BASE} & 79.24 & 89.73 & 79.64 & 77.98 & 81.73 & 82.1 & 83.22 & 82.41 & 82.01 \\
    & LiLT[InfoXLM]\textsubscript{BASE} & 85.74 & 90.47 & 80.88 & 83.40 & 85.77 & 87.92 & 87.69 & 84.93 & 85.85 \\
    & XFormParser &\textbf{93.89} & \textbf{94.02} & \textbf{90.94} & \textbf{90.19} & \textbf{89.72} & \textbf{91.74} & \textbf{91.94} & \textbf{90.94}  &\textbf{91.67}\textsubscript{{\color{red} (+5.82)}} \\
    \cmidrule{1-11}
    \multirow{5}*{RE}
    & XLM-RoBERTa\textsubscript{BASE} & 36.38 & 67.97 & 68.29 & 68.28 & 67.27 & 69.37 & 68.87 & 60.82 & 63.41 \\
    & InfoXLM\textsubscript{BASE} & 36.99 & 64.93 & 64.73 & 68.28 & 68.31 & 66.90 & 63.84 & 57.63 & 61.45 \\
    & LayoutXLM\textsubscript{BASE} & 66.71 & 82.41 & 81.42 & 81.04 & 82.21 & 83.10 & 78.54 & 70.44 & 78.23 \\
    & LiLT[InfoXLM]\textsubscript{BASE} & 74.07 & 84.71 & 83.45 & 83.35 & 84.66 & 84.58 & 78.78 & 76.43 & 81.25 \\
    & XFormParser & \textbf{97.00} & \textbf{95.49} & \textbf{94.53} & \textbf{95.67} & \textbf{96.76} & \textbf{97.3} & \textbf{95.49} & \textbf{95.06}&\textbf{95.89} \textsubscript{{\color{red} (+14.64)}}\\
    \bottomrule
    \end{tabular}
    \caption{Multi-language fine-tuning accuracy (F1) on the XFUND dataset (fine-tuning on 8 languages all, testing on X), where “SER” denotes the semantic entity recognition and “RE” denotes the relation extraction.}
    \label{tab:result_xfund_multi}
\end{table*}

\section{Experiments and Discussion}

\subsection{Experiment Settings}

\textbf{Datasets.}
FUNSD~\cite{jaume2019} is a scanned document dataset for form parsing. It is a subset of the RVL-CDIP~\cite{harley2015evaluation} and consists of 149 training samples and 50 test samples with various layouts. XFUND \cite{xu-etal-2022-xfund} is a multilingual form parsing benchmark. It includes 1,393 fully annotated forms in 7 languages. Each language contains 199 forms, with 149 forms in the training set and 50 forms in the test set.

XFUND and FUNSD datasets have two tasks: SER, where BIO labeling format is used for sequence labeling of each entity, and RE, where all possible pairs of given semantic entities are generated to gradually construct a candidate set of relations, identifying all entity pairs with existing associations.

In this paper, InDFormSFT is constructed based on the Chinese-English context and eight application scenarios, the training set contains 422 samples, and the validation set and test set contain 70 samples each. Compared with FUNSD and XFUND datasets, larger datasets can provide more samples for the model to learn and train, which helps to improve the generalization ability and performance of the model. Please refer to Appendix~\ref{sec:indformsft} for the construction process of InDFormSFT.




\begin{table*}[!h]
    \fontsize{8}{8}\selectfont
    \centering
    \begin{tabular}{c|cccccccccc} 
    \toprule

    &Model&FUNSD&ZH&JA&ES&FR&IT&DE&PT&Avg.\\
    \midrule
    \multirow{5}*{SER} &XLM-RoBERTa\textsubscript{BASE}&66.70&41.44&30.23&30.55&37.10&27.67&32.86&39.36&38.24\\
&InfoXLM\textsubscript{BASE}&68.52&44.08&36.03&31.02&40.21&28.80&35.87&45.02&41.19\\
&LayoutXLM\textsubscript{BASE}&79.40&60.19&47.15&45.65&57.57&48.46&52.52&53.90&55.61\\
&LiLT[InfoXLM]\textsubscript{BASE}&84.15&61.52&51.84&51.01&59.23&53.71&60.13&63.25&60.61\\
& XFormParser & \textbf{92.46} & \textbf{72.00} & \textbf{58.25} & \textbf{59.12} & \textbf{69.92} & \textbf{73.72} & \textbf{71.88} & \textbf{73.46} & \textbf{71.35}\textsubscript{{\color{red} (+10.74)}}  \\
    \midrule
    \multirow{6}*{RE}
    &XLM-RoBERTa\textsubscript{BASE}&26.59&16.01&26.11&24.40&22.40&23.74&22.88&19.96&22.76\\
    &InfoXLM\textsubscript{BASE}&29.20&24.05&28.51&24.81&24.54&21.93&20.27&20.49&24.23\\
    &LayoutXLM\textsubscript{BASE}&54.83&44.94&44.08&47.08&44.16&40.90&38.20&36.85&43.88\\
    &LiLT[InfoXLM]\textsubscript{BASE}&62.76&47.64&50.81&49.68&52.09&46.97&41.69&42.72&49.30\\
    &GOSE[LiLT]&76.97&69.30&68.05&70.72&71.45&63.55&59.97&58.30&67.29\\
    & XFormParser & \textbf{91.24} & \textbf{74.02} & \textbf{81.77} & \textbf{83.02} & \textbf{79.60} & \textbf{78.61} & \textbf{80.18} & \textbf{81.01} & \textbf{81.18} \textsubscript{{\color{red} (+13.89)}}\\  
    \bottomrule
    \end{tabular}
    \caption{Cross-lingual zero-shot transfer F1 accuracy on FUNSD and XFUND (fine-tuning on FUNSD, testing on XFUND).}
    \label{tab:result_xfund_zero-shot}
\end{table*}


\subsection{Language-specific Fine-tuning}
Table~\ref{tab:result_xfund_specific} presents the results of specific language fine-tuning experiments on the XFUND and FUNSD public datasets. Each column in the table represents a different language (FUNSD, ZH, JA, ES, FR, IT, DE, PT) \cite{xu-etal-2022-xfund}. The table compares different models, including XLM-RoBERTa\textsubscript{BASE}, InfoXLM\textsubscript{BASE}, LayoutXLM\textsubscript{BASE}, LayoutLMv3\textsubscript{BASE}, LiLT[InfoXLM]\textsubscript{BASE}, GeoLayoutLM, and XFormParser\textsubscript{BASE}. The numerical values in the table indicate the performance metrics of each model fine-tuned in specific languages. XFormParser demonstrates superior performance compared to other models in most languages. On the FUNSD dataset, XFormParser achieves a performance of 92.46\%. XFormParser also exhibits strong performance on the XFUND datasets in different languages, with average performance metrics of 89.04\% (SER) and 90.54\% (RE).

\begin{table*}[!htbp]
    \fontsize{10}{10}\selectfont
    \centering
    \begin{tabular}{clccccccccc} 
    \hline
    &\multirow{2}*{Method}&\multicolumn{2}{c}{Task}&\multicolumn{2}{c}{Component}&\multicolumn{2}{c}{SER F1 Accuracy}&\multicolumn{2}{c}{RE F1 Accuracy↑} \\
    \cmidrule(lr){3-4}
    \cmidrule(lr){5-6}
    \cmidrule(lr){7-8}
    \cmidrule(lr){9-10}
    && SER & RE & Decoder & soft label & EN & ZH & EN & ZH \\
    \hline
    & XFormParser & \cmark{} &  \cmark{} & \cmark{} & \cmark{} & \textbf{92.46} & \textbf{93.42} & \textbf{91.2} & \textbf{93.14} \\
    \hline
    1&w/o Task RE & \cmark{} &  \xmark{} & \xmark{} & \xmark{} & 91.89 & 92.86  & - & - \\
    2&w/o Task SER & \xmark{} &  \cmark{} & \cmark{} & \cmark{} & - & -  & 90.90 & 92.64\\
    3&w/o Decoder & \cmark{} &  \cmark{} & \xmark{} & \cmark{} & 91.40 & 92.75 & 79.79 & 81.73\\ 
    4&w/o soft label & \cmark{} &  \cmark{} &  \cmark{} & \xmark{} & 91.19 & 92.19 & 90.75 & 91.20\\
    \hline
    \end{tabular}
    \caption{Ablation study of our model using LayoutXLM as the backbone, InDFormSFT
as training dataset and test on the FUNSD and XFUND (ZH). The symbol EN denotes FUNSD and ZH means Chinese language.}
    \label{tab:result_xfund_ablation}
\end{table*}

\subsection{Multi-language Fine-tuning}
XFUND dataset inspired Multilingual fine-tuning Task, refers to multilingual task fine-tuning on XFUND dataset, all trained on 8 languages and tested on a specific language. 
As shown in Table~\ref{tab:result_xfund_multi}, in the multilingual training task, XLM-RoBERTa\textsubscript{BASE}, InfoXLM\textsubscript{BASE}, LayoutXLM\textsubscript{BASE}, and XFormParser\textsubscript{BASE} all show stronger ability than language-specific fine-tuning, obtaining higher F1 score. In the case of multiple languages, XFormParser\textsubscript{BASE} shows a more powerful extraction ability. XFormParser performs best in the multi-language fine-tuning experiment. The average F1 accuracy of SER task is 91.67\%. For RE task, XFormParser also achieves the best performance in multi-language fine-tuning experiments, with an average F1 of 95.89\%. These results underscore the efficacy of the XFormParser in handling multi-language tasks within the XFUND dataset, particularly in comparison to other baseline and advanced models. With more data, our model demonstrates stronger learning capabilities.

\subsection{Zero-shot Fine-tuning}
Previous experiments illustrate that our method achieves improvements using full training samples. We explored the transferability of our model structure and found that it will gain strong transferability on RE tasks. Thus, we compare with the previous SOTA model LiLT on few-shot settings. The experimental results in Table~\ref{tab:result_xfund_zero-shot} indicate that the average performance of XFormParser still outperforms the SOTA model LiLT and GOSE. In the task of SER, XFormParser exhibited the highest F1 scores across all languages, achieving an average of 71.35\%, with notable improvements in Chinese (72.00\%) and Portuguese (73.46\%).
In the task of RE, the highest performance in the RE task was again seen with XFormParser, which achieved an impressive average F1 score of 81.18\%, showing its robustness in relation extraction across different languages.
the XFormParser consistently outperforms other models in both the SER and RE tasks across various languages, highlighting its superior cross-lingual transfer capabilities, and it can improve the generalization of the model.

\subsection{Ablation Study}

\begin{table}[!htbp]
\centering
\setlength{\tabcolsep}{3mm}{ 
\begin{tabular}{ccc}
\toprule
             epoch start&epoch warm&RE F1 \\
             \midrule
             \xmark{}&\xmark{}&91.20\\ 
             10&\cmark{}&92.67\\
             20&\cmark{}&92.74\\
             30&\xmark{}&92.44\\
             30&\cmark{}&93.14\\
             40&\xmark{}&91.06\\
             40&\cmark{}&91.35\\
             50&\xmark{}&90.93\\
             50&\cmark{}&91.43\\
\bottomrule
\end{tabular}}
\caption{For the ablation experiments of epoch start and epoch warm, epoch start refers to the epoch round when the soft label is started, epoch warm refers to whether to use the transition mechanism, and the total training rounds are 100. The first line refers to the method of warm-up soft label that is not applicable.}
\label{tab:Ablation_soft label}
\end{table}

To better understand the working principle of the model and determine the extent to which the key components or strategies contribute to the model performance, we designed ablation experiments to verify the feasibility of the method proposed in this paper. In ablation experiments, a series of modifications or eliminations are made to the original model to see how these modifications affect the model's performance. Ablation experiments are designed and implemented from three innovations of our method:

\textbf{Effectiveness of Individual Components.} We further investigate the effectiveness of different modules in our method. we compare our model with the following variants in Table~\ref{tab:result_xfund_ablation}.

1) w/o multi-task. In this variant, we try to remove the multi-task training method and only use SER or RE for training. This change causes a significant performance decay. The results shown in Table~\ref{tab:result_xfund_ablation} suggest that training two tasks at the same time has the same effect on both SER and RE tasks. With improvement, two tasks that share PTM parameters will learn cross-information that improves this task.

2) w/o Decoder. In this variant, we remove the decoder. This change causes a significant performance decay. This suggests the injection of an additional decoder can guide the powerful decoding capabilities of entities and provide strong dependencies for relation classification.

3) w/o Warm-up soft label. 
In this variant, we remove the training method Warm-up soft label from XFormParser. This change means the model only uses hard labels due to the training. The results shown in Table~\ref{tab:result_xfund_ablation} indicate that a Warm-up soft label can improve the effect of the model and prevent the model from overfitting.
Experimental data Table~\ref{tab:Ablation_soft label} indicates that as epoch starts increases, model performance initially improves but then decreases. This trend suggests that the model learns sufficient data features by a certain stage, and further training with a Warm-up soft label does not enhance performance and may even lead to overfitting.
\subsection{Visualization Display}
\begin{figure}[!h] 
	\centering
	\subfloat[]{
		\includegraphics[width=1.0\columnwidth]{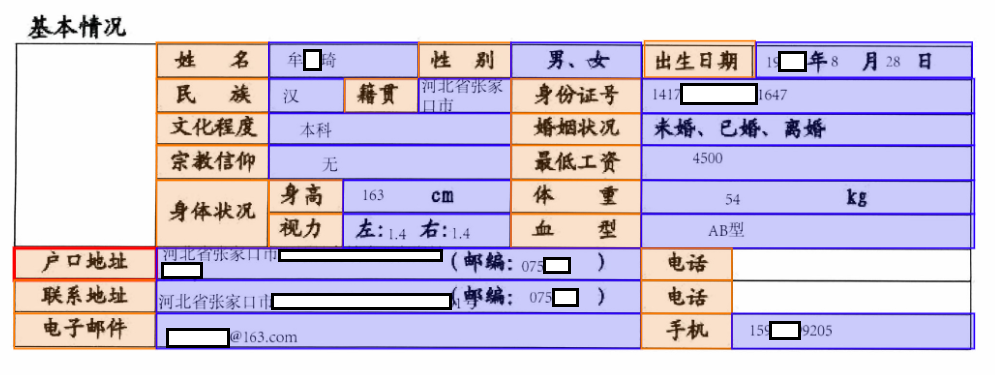}
		\label{subfig:cswin}
	}\\
	\subfloat[]{
		\includegraphics[width=1.0\columnwidth]{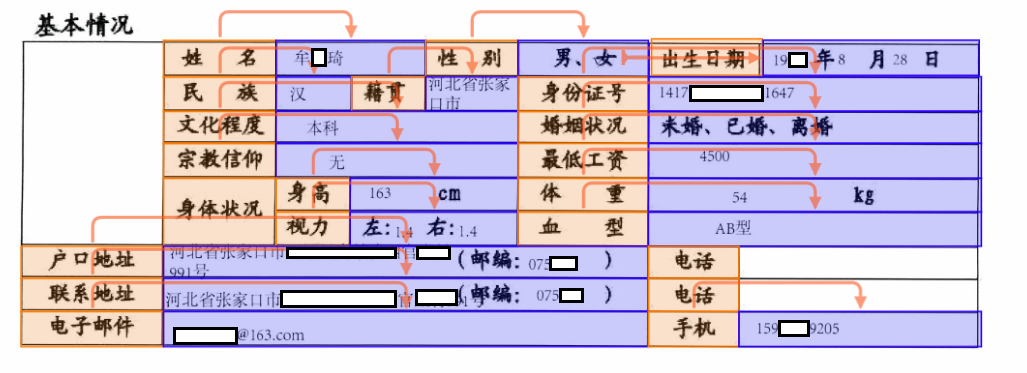}
		\label{subfig:mhsa}
	}
	
	\caption{Illustration of (a) The form image that is entered into the system; (b) Visualization of SER and RE results.}
	\label{fig:result}
\end{figure}

We visualize the SER and RE of XFormParser on a text-intensive form image as shown in Figure \ref{fig:result}(b), where the orange boxes are named entities and the arrows represent the matching relations between the entities. This figure confirms the effectiveness of XFormParser.


\section{Conclusion}
Aiming at the common problems in rich text form parsing tasks, this paper proposes a semi-structured form parser XFormParser based on multi-modal and multi-lingual knowledge. XFormParser integrates layoutXLM pre-trained backbone, semantic entity recognizer, and relation extraction decoder, and implements SER and RE tasks for semi-structured form parsing. At the same time, to enrich the experimental data in this field and improve the parsing ability of industrial application scenarios, this paper constructs a Chinese and English multi-scenario form parsing SFT dataset InDFormSFT. Four different Settings (such as Language-specific fine-tuning, Multi-language fine-tuning, Cross-lingual fine-tuning, and Zero-shot) are designed on two benchmark datasets and InDFormSFT ), and the results show the effectiveness and superiority of XFormParser.

\section{Limitations}
The purpose of this work is to provide a simple, efficient, and easy-to-deploy semi-structured form parsing component for the end side (PC or mobile). Although we use a multi-modal approach and expand the training set to improve the performance of the model, while taking into account the multi-language parsing scenario, this work still has the following limitations.
\paragraph{Diversity of Languages.} InDFormSFT only includes two languages, Chinese and English, and lacks expansion of form knowledge for the other six languages in XFUND. After that, multi-language augmented data can be constructed by using the same data set construction process with the help of machine translation and layout design algorithms.
\paragraph{Model Diversity.} The comparative experiments do not include the experimental results of the large model of multi-modal documents on relevant benchmarks, thus lacking the most powerful demonstration of the upper bound of the performance of the model for the current task. In addition, our work does not further improve the multilingual pre-trained model backbone and directly adopts the strongest and most easy-to-use model investigated as the backbone of XFormParser. These limitations need to be further studied and improved.
\paragraph{Completeness of Verification.} There is a lack of validation of model compression and inference acceleration methods such as model distillation, model pruning, and model quantization.

\section*{Acknowledgments}
This work was partially supported by the National Natural Science Foundation of China (Grant Nos. 62276017, 62406033,  U1636211, 61672081, 62272025), and the State Key Laboratory of Complex\& Critical Software Environment (Grant No. SKLCCSE-2024ZX-18).

\bibliography{custom}

\appendix

\section{Build the Warm-up Soft Label}
\label{sec:softlabel}
Soft labeling, also known as soft targets or probabilistic labeling, is a technique in natural language processing where labels for training data are represented as probability distributions rather than as hard, binary labels. Soft labels provide more information about the uncertainty and distribution of classes, leading to better generalization. In traditional hard labels, each sample can only belong to one category, while using soft labels can represent situations where a sample may belong to multiple categories.

We used the improved warm-up soft label, which can effectively improve the performance when applied to fine-tuning form parsing models. In the beginning stage of training, hard labels are used to supervise the model so that it can converge quickly. In the middle and later stages of training, soft labels are used to help model training. During the transition period between soft and hard labels, a warming mechanism is added, and the weight of soft tags is continuously increased. This is useful for tasks such as multi-label classification. 
For the soft labels $logits_{ser}$ and its embedding are calculated. The specific formula is as follows,
\begin{equation}
     LE_{sl} = \frac{softmax(logits_{ser})\cdot LE_{weight}}{N}
\end{equation}
where $N$ is the number of tags, $LE_{weight}$ is the weight of the word list that stores tag embeddings, $logits_{ser}$ first obtains the distribution probability through a layer of softmax, and $softmax(logits_{ser})$ Perform dot multiplication with $LE_{weight}$ to get a weighted label embedding vector. Finally, the weighted label embedding vector is divided by the number of samples in the dataset $N$ to obtain the average value of the label embedding vector $LE_{soft label}$, $LE_{sl}$ for short.

\begin{figure}[!t]
	\centering
	\includegraphics[width=1.0\linewidth]{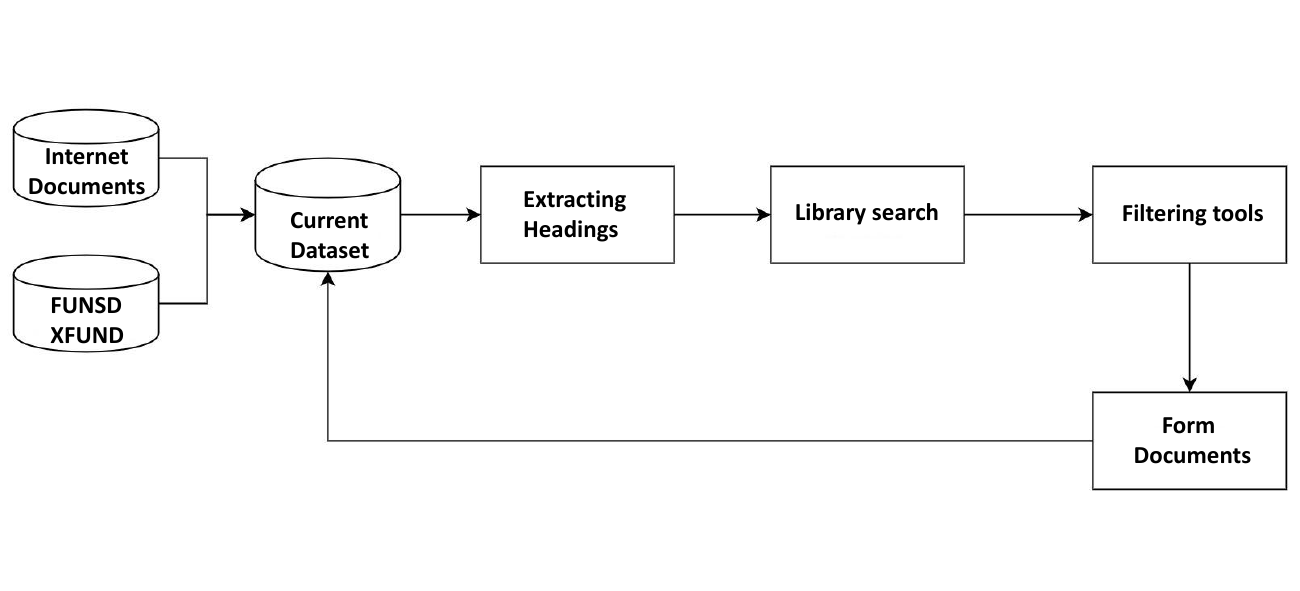}
	\caption{It shows the process of data search. On the basis of the constructed data, the title of the document is extracted as the search term, and other form files are searched in the document search engine.}
	\label{fig:souji}
\end{figure}
\begin{figure}[!t]
	\centering
	\includegraphics[width=1.0\linewidth]{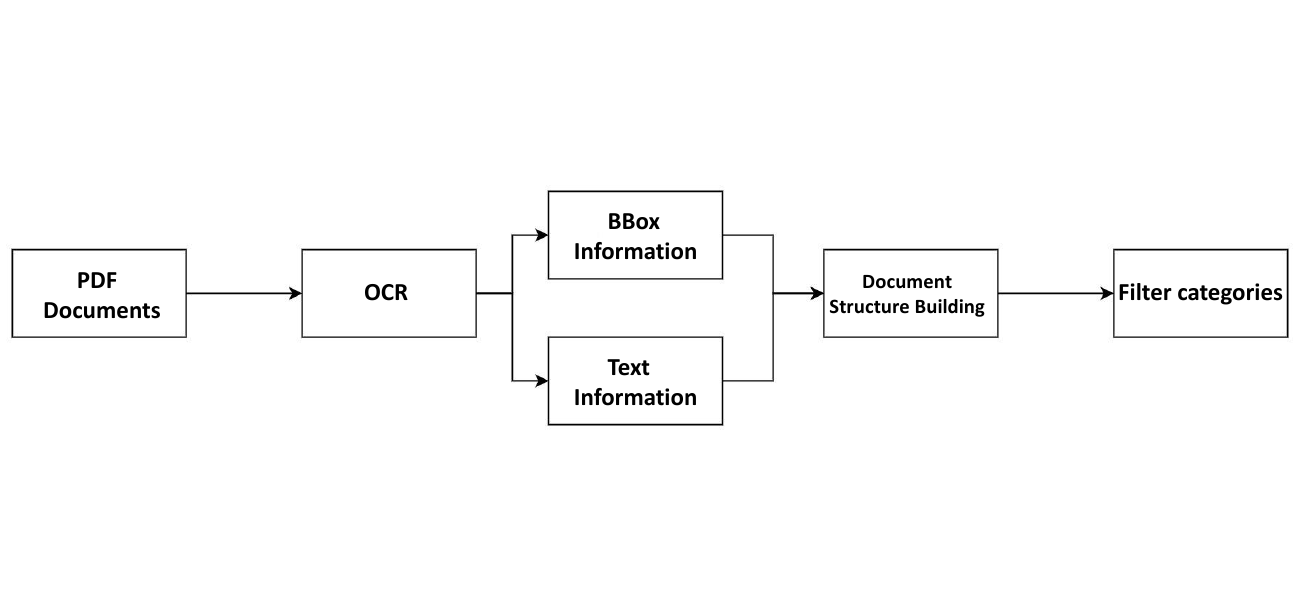}
	\caption{Firstly, the optical character recognition tool is used to process the file, and the text information and border information are obtained. The data structure of the form is constructed through the border and text, including the structure information and the text information of the form.}
	\label{fig:shaixuan}
\end{figure}
\begin{figure}[!t]
	\centering
	\includegraphics[width=1.0\linewidth]{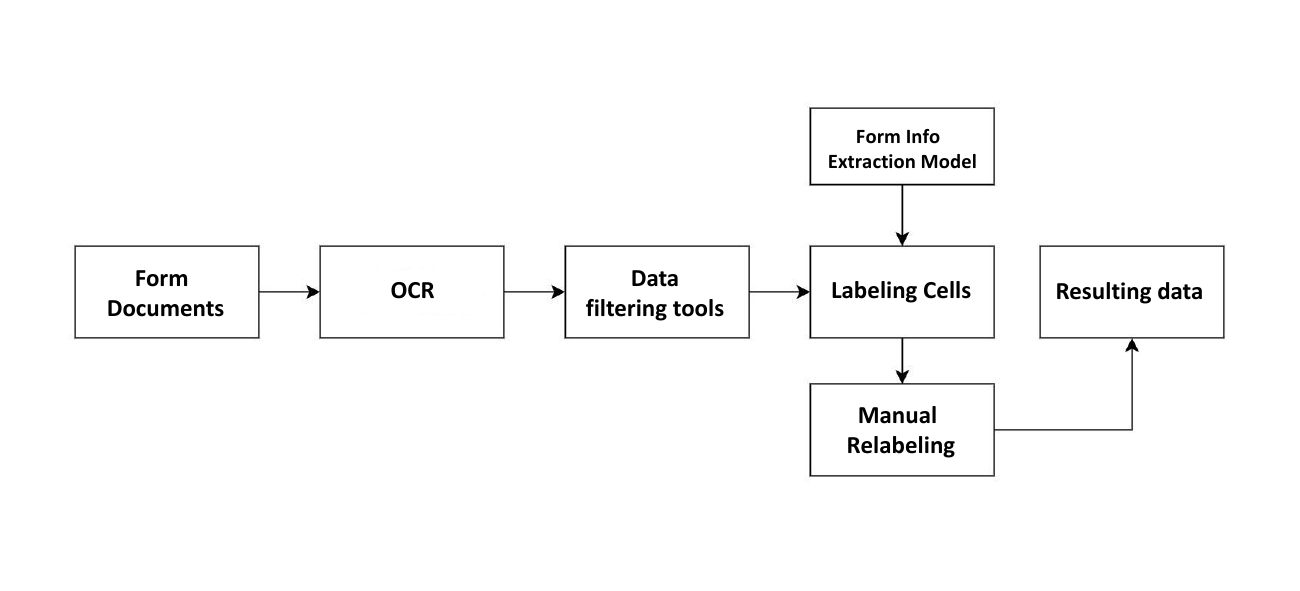}
	\caption{It shows the construction of a form data auxiliary labeling tool. Firstly, the optical character recognition tool is used to process the file to obtain text and border information. Then the form data filtering tool is used to determine whether the cell can be labeled. Then the form information extraction model is used for auxiliary labeling, and finally, the data result is obtained by manual labeling.}
	\label{fig:biaoji}
\end{figure}

\begin{figure*}[!t]
	\centering
	\includegraphics[width=1.0\linewidth]{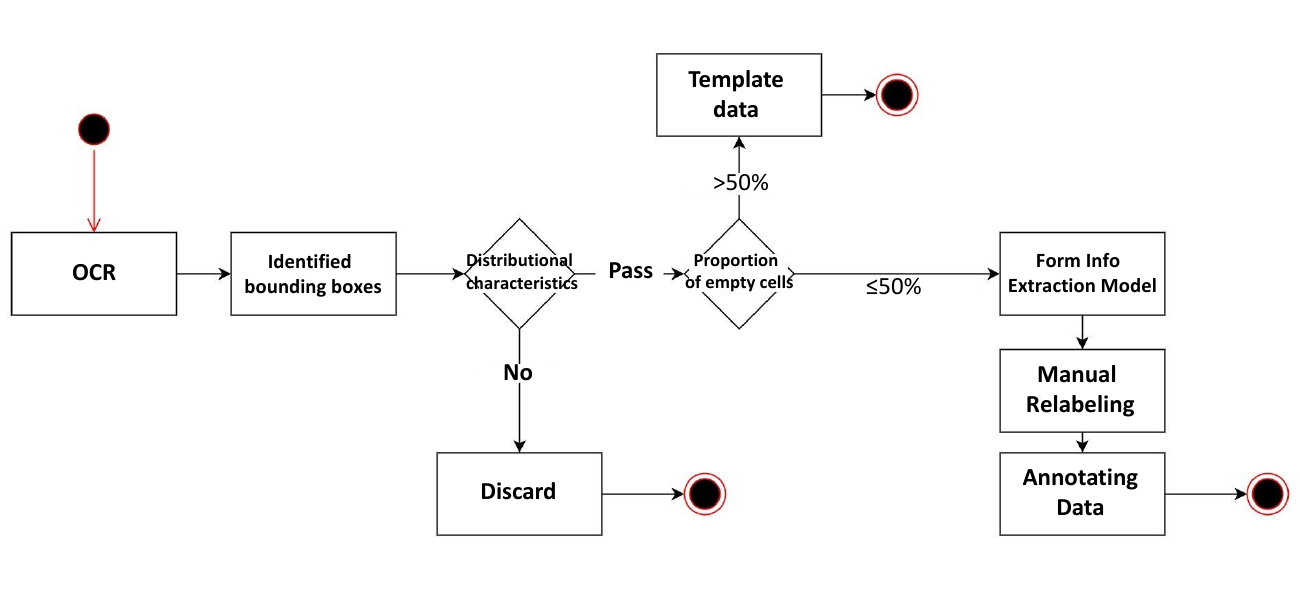}
	\caption{Firstly, a PDF form is obtained, the text and border information are extracted through OCR, and the cell characteristics are calculated to determine whether it conforms to the form structure. If it does not meet the conditions, it is directly discarded. If it meets the conditions, the vacancy rate of the cell is calculated. The qualified forms are pre-predicted and labeled by the model, and the pre-labeled results are imported into the labeling system. Finally, the correct labeled data are obtained by manual verification for training and testing.}
	\label{fig:labeling}
\end{figure*}

In actual experiments, it was found that the training effect was not good in the first half of model training. Although soft labels contain more information, they cannot guide the model in learning tasks in the early stage of training, so hard labels are still used in the early stage of model training. After the model has been trained to have preliminary capabilities, then use soft label training, and provide a transition for the conversion of hard label and soft label. 
In fact, the final label embedding calculation method is as follows:
\begin{equation}
     \alpha = \min(1,(ep-ep_{start}/ep_{warm}))
\end{equation}
\begin{equation}
LE = \begin{cases}
   LE_{hl}, & ep \leq ep_{start} \\
   \alpha LE_{sl} + \beta LE_{hl}, & ep > ep_{start} \\
\end{cases}
\end{equation}
where $\alpha$ is a parameter that decays sublinearly with training and $\beta$=$1$-$\alpha$. where $ep$ represents the current training epoch, $ep_{start}$ denotes the starting epoch, and $ep_{warm}$ is a predefined value determining the warm-up duration. The parameter $\alpha$  acts as a scaling factor for the learning rate, ensuring a gradual transition from an initial learning rate to the desired rate, as the training progresses.

\section{Approach to building InDFormSFT}
\label{sec:indformsft}
The large labeled data set is the main support for the high performance of deep learning. Form datasets are a common type of datasets used to collect, store, and analyze user-submitted data. It is commonly used in various application areas such as market research, user surveys, online registration, order forms, etc. In recent years, there have been some academic data sets in the field of forms. This kind of data set contains two kinds of information, one is the image of the original document, and the other is the specific information of the document that has been annotated or parsed, usually including the image, the coordinates of the text box, the text content of the text box, the label of the text box and the relationship between the text box.
\subsection{Data Collection and Annotation}
The basic data set is constructed through the Chinese and English data sets of FUNSD and XFUND, and then the Chinese form data on the Internet is collected. To avoid privacy and sensitive information issues of real-world documents, this paper collects documents publicly available on the Internet. Semi-structured forms were collected through Baidu Library, and the search keywords included: a comprehensive table of university teachers, a comprehensive table of senior safety engineers, a comprehensive table of professor-level senior engineers, etc. By downloading and collecting the files of the Shenzhen Stock Exchange, Shanghai Stock Exchange, and other financial platforms, the form files that meet the task definition are found. 

Figures~\ref{fig:souji},~\ref{fig:shaixuan},~\ref{fig:biaoji} show the data search process, the form data filtering, and the construction process of the auxiliary labeling tool, respectively. Finally, to facilitate the pre-processing and labeling of forms, a set of engineering developments of data screening and labeling process based on Chinese forms was completed, as shown in Figure~\ref{fig:labeling} below.

\subsection{Instances of Semi-structured Data}
The form data also aligns with the format of the XFUND dataset, with cell granularity, where each cell contains information such as absolute cell coordinates (box), cell text information (text), cell label information (label), cell ids (id), and linking between cells (linking).

\begin{figure}[!h] 
	\centering
	\subfloat[]{
		\includegraphics[width=1.0\columnwidth]{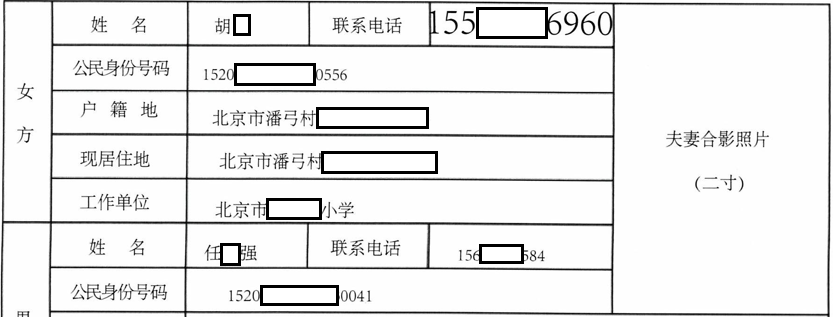}
		\label{subfig:tuli}
	}\\
	\subfloat[]{
		\includegraphics[width=1.0\columnwidth]{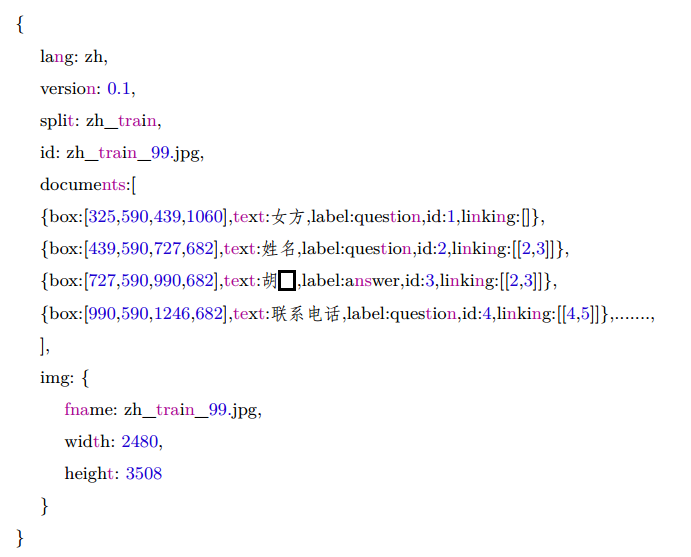}
		\label{subfig:json}
	}
	
	\caption{Illustration of (a) A form image in InDFormSFT; (b) The annotation corresponding to the first row of the image.}
	\label{fig:labelimg}
\end{figure}

The first row of the picture is shown in Figure~\ref{fig:labelimg}(a), and the corresponding annotated data format is shown in Figure~\ref{fig:labelimg}(b).
\begin{table*}[!ht]
    \fontsize{8}{8}\selectfont
    \centering
    \begin{tabular}{c|cccccc} 
    \toprule

    Dataset partition &Data volume & Question Entity & Answer Entity &Single Entity & Title Entity & Continuous char entity\\
    \midrule
    Training set &422 &6702 &6825 &422 &370 &194\\
    Validation set &70 &1375 &1443 &65 &18 &83\\
    Testing set & 70 & 1468 & 1644 &78 &58 &18 \\
    \bottomrule
    \end{tabular}
    \caption{Analysis of the number of entity labels in InDFormSFT.}
    \label{tab:entityfx}
\end{table*}
\begin{table*}[!h]
    \fontsize{9}{9}\selectfont
    \centering
    \begin{tabular}{c|cccc} 
    \toprule

    Dataset partition &One-to-one & One-to-two & One-to-three &One-to-many \\
    \midrule
    Training set &11806 &265 &96 &171 \\
    Validation set &2366 &55 &38 &39 \\
    Testing set &2626 &76 &53 &72 \\
    \bottomrule
    \end{tabular}
    \caption{Entity relationship analysis of InDFormSFT.}
    \label{tab:relationfx}
\end{table*}
In Figure~\ref{fig:labelimg}(b), lang denotes the language of the text. In this case, its value is "zh", indicating that the text is Chinese. version indicates the version of the dataset, where the value is "0.1" and "split", which means the dataset is split into train, val, and test sets. "id" represents a unique identifier for the form data. documents contain a list of form cell information, and box represents the coordinates of the text's bounding box on the image. The value is a list of four integers representing the top-left x-coordinate, the top-left y-coordinate, the bottom-right x-coordinate, and the bottom-right y-coordinate. text represents the text content of the cell. Here, the value is a string that contains some text. label denotes the label content of the cell, which includes the SINGLE entity (SINGLE), QUESTION entity (QUESTION), ANSWER entity (ANSWER), and continuous character entity (ANSWERNUM) mentioned above. id represents the unique identifier of the cell. linking represents a linking relationship between cells and is a list of entity pairs with two ids, the first for the question entity and the second for the answer or consecutive character entity. img contains the form image information, where fname represents the name of the image file, width represents the width of the image, and height represents the height of the image.

\subsection{Analysis of InDFormSFT}

As shown in Table~\ref{tab:entityfx}, for the analysis of the entity content of the data set, according to the table content analysis, the entity labels of the data set of this paper have the following characteristics: the number of question entities and answer entities is large. In the training set, the number of question entities is 6702, the number of answer entities is 6825, and the number of question entities and answer entities is basically the same. This indicates that there are a large number of question-and-answer entities in the dataset that need to be identified and annotated. These entities may include person names, place names, organizations, etc., and we need our model to be able to accurately identify and label these entities. The number of single entities, title entities, and continuous character entities is relatively rare, and recognizing these sparse entities is a challenging task.

According to the content analysis of Table~\ref{tab:relationfx}, the table shows the division of the data set and the corresponding relationship types. Entities are classified by relationship type, including one-to-one, one-to-two, one-to-three and one-to-many (greater than three). These relation types describe the number of entity-time correspondences in the form information extraction task. We can see that most relationships are concentrated in one-to-one relationships, and a few exist in one-to-two, one-to-three, and one-to-many (greater than three) relationships. By analyzing the distribution of the number of different correspondences, we can get the distribution of the number of samples of different relation types in the dataset.

\subsection{Evaluation Metrics}
\label{app:b4}
In this paper, the target tasks for the datasets used are divided into SER and RE. For the SER task, the model needs to determine the class of each Cell in the form: SINGLE, QUESTION, ANSWER, and ANSWERNUM. The evaluation metric is Cell Acc (CA). The accuracy of Cell Discrimination is determined by Correct Cell Discrimination (CCD) and Total Cell Count (TCC). The formula is as follows:
\begin{equation}
     CA=\frac{CCD}{TCC}
\end{equation}
For the cell relation linking task, we need to determine which combinations of cells in each table have a key-value relation. The F1-Score is a commonly used metric to evaluate the performance of classification models, which takes into account both Precision and Recall. The formula for calculating the F1 score is as follows:
\begin{equation}
     F1=2\times\frac{Precision\times Recall}{Precision+Recall}
\end{equation}
\begin{equation}
     Precision=\frac{TP}{TP+FP}
\end{equation}
\begin{equation}
     Recall=\frac{TP}{TP+FN}
\end{equation}
In this task, TP represents the number of entity pairs correctly predicted by the model as having a relationship. Actually, having a relationship, FP represents the number of entity pairs predicted by the model as having a relationship but actually having no relationship. FN represents the number of entity pairs predicted by the model as having no relationship but actually having a relationship. The F1 value ranges from 0 to 1, with values closer to 1 indicating better performance of the model.
\end{document}